\input ./style/arxiv-general.cfg
\documentclass[aoas,MSNbibl,nameyear,seceqn,noautosecdot,dvips]{arximspdf}
\makeatletter
   \@ifpackageloaded{graphicx}{}{\usepackage{graphicx}}
\makeatother

\doi{10.1214/15-AOAS812}
\volume{9}
\issue{2}
\pubyear{2015}
\firstpage{801}
\lastpage{820}
\docsubty{FLA}

\makeatletter
\newproclaim{ex}{Example}
\newcommand{\eqref}[1]{(\ref{#1})}
\newcommand{\hw}{\hat{w}}
\newcommand{\hz}{\hat{z}}
\newcommand{\hmu}{\hat{\mu}}
\newcommand{\hrho}{\hat{\rho}}
\newcommand{\nn}{\mathcal{N}}
\newcommand{\EEe}[2][]{\mathbb{E}_{#1}\bigl[#2\bigr]}
\newcommand{\PPp}[2][]{\mathbb{P}_{#1}[#2]}
\newcommand{\argmin}{\operatorname{argmin}}
\newcommand{\HE}{\mathrm{HE}}
\newcommand{\WH}{\mathrm{WH}}
\newcommand{\law}{\mathcal{L}}
\makeatother

\begin{document}
\begin{frontmatter}

\title{Weakly supervised clustering: Learning fine-grained signals
from coarse labels}
\runtitle{Weakly supervised clustering}

\begin{aug}
\author[A]{\fnms{Stefan}~\snm{Wager}\corref{}\thanksref{M1,T1}\ead[label=e1]{swager@stanford.edu}},
\author[B]{\fnms{Alexander}~\snm{Blocker}\thanksref{M2}\ead[label=e2]{awblocker@google.com}}
\and
\author[B]{\fnms{Niall}~\snm{Cardin}\thanksref{M2}\ead[label=e3]{niallc@google.com}}
\runauthor{S. Wager, A. Blocker and N. Cardin}
\affiliation{Stanford University\thanksmark{M1} and Google,
Inc.\thanksmark{M2}}
\address[A]{S. Wager\\
Department of Statistics\\
Stanford University\\
Stanford, California 94305\\
USA\\
\printead{e1}}
\address[B]{A. Blocker\\
N. Cardin\\
Google, Inc.\\
Mountain View, California 94043\\
USA\\
\printead{e2}\\
\phantom{E-mail: }\printead*{e3}}
\end{aug}
\thankstext{T1}{Supported by a B. C. and E. J. Eaves Stanford Graduate Fellowship.}

%
\received{\smonth{6} \syear{2014}}
%
\revised{\smonth{2} \syear{2015}}

%
\begin{abstract}
Consider a classification problem where we do not have access to labels
for individual
training examples, but only have average labels over subpopulations. We
give practical examples of this setup and show how such a
classification task can usefully be analyzed as a \emph{weakly
supervised clustering problem}. We propose three approaches to solving
the weakly supervised clustering problem, including a latent variables
model that performs well in our experiments. We illustrate our methods
on an analysis of aggregated elections data and an industry data set
that was the original motivation for this research.
\end{abstract}

%
\begin{keyword}
\kwd{Latent variables model}
\kwd{uncertain class label}
\end{keyword}
\end{frontmatter}

\setcounter{footnote}{1}

\section{\texorpdfstring{Introduction.}{Introduction}}

A search provider wants to know whether people who clicked on a given
search result found it useful.\footnote{This example is hypothetical,
but conveys the key difficulties from a real problem faced by a large
internet company.}
A searcher's behavior can provide valuable clues as to whether she
liked the result: if she immediately hit the back button upon seeing
the landing page, she probably had a bad experience. Conversely, a
searcher interacting with the result may be seen as a positive signal.

Many online providers seek to directly estimate user happiness with
click-level proxies. For example, in the context of web search, one
well-known signal of user dissatisfaction is a ``bounce,'' where people
go to a search result but then immediately return to the search page
[\citet{levy2011plex}, page 47, \citet{sculley2009predicting}].
\citet{bucklin2009click} give an overview of how data about site
usage patterns is used in online marketing.
However, using hand-crafted proxies to understand user experience has
its limits. It requires analysts to map these proxies to user
satisfaction in a usually unprincipled way, and different proxies may
lead to contradicting conclusions.

This paper addresses the question: how can we combine multiple
click-level features into a single principled measure of user
satisfaction? The main difficulty is that we have no explicit response
to train on, as searchers do not tell us whether or not they were
satisfied with any given click.
What we do have is side information about whether some subpopulation of
clicks was mostly satisfied or not: in the context of our example, we
might know from outside sources (e.g., human raters) that some search
results are good ones and that most users who click to them should be
satisfied, whereas other results are of lesser quality and may leave
some searchers disappointed.

Formally, we are faced with a binary classification task where we do
not have labels for individual clicks, but only have a rough idea of
the average fraction of satisfied clicks over large subpopulations. In
other words, we have a classification task where the available training
labels are much coarser-grained than the signal we want to fit.

We adopt a weakly supervised approach, where we use the coarse training
labels to guide a clustering algorithm. At a fundamental level, we
expect satisfied versus unsatisfied behaviors to look different from
each other in a way that does not depend on group (here, the search
result); thus, we should be able to construct a global clustering of
clicks that respects this distinction. But there are presumably many
natural ways to divide clicks into two groups other than the
satisfied/unsatisfied distinction: we might expect energetic/tired or
hurried/leisurely clicks to also split into distinct clusters. Our goal
is to use side information to avoid this issue and pick out the
``right'' way of clustering the data. We do this by forcing the
clustering algorithm to respect marginal class memberships for
different subpopulations: concretely, we want most clicks on good
search results to be in the good cluster, whereas clicks on the
mediocre results should be more evenly split.

We call this task of finding a clustering of the data that respects
side information about marginal cluster membership for multiple
subpopulations a \emph{weakly supervised clustering problem}. This
problem surfaces when we want to understand click or behavior level
data, but only have access to coarse-grained side information for
training. Other examples that can be cast as weakly supervised
clustering problems include the following.


\begin{ex}\label{ex1}
An online advertiser wants to understand what
kind of click-level interaction with an ad suggests that a customer
will later visit a physical store. It is not always practical to ask
users directly whether or not they visited a store after seeing an ad,
and so this is not a standard supervised problem. However, the
advertiser may have some idea about how successful the ads were at a
campaign level. With a weakly supervised clustering approach, it can
use this highly aggregated campaign-level signal to learn how to
interpret click-level behaviors.
\end{ex}

\begin{ex}\label{ex2}
A political scientist wants to study how
different demographic groups voted in an election. However, instead of
having access to voter-level data, she only gets to see aggregated
state-level election data. In Section~\ref{secvoteexample}, we cast
this example as a weakly supervised clustering problem and use our
method to analyze aggregated data from a US presidential election.
\end{ex}

In this paper, we compare three possible approaches to the weakly
supervised clustering problem: a latent variables model, a method of
moments estimate, and a naive approach that turns the problem into a
supervised problem using a hard assignment. We find the method of
moments approach to be prohibitively unstable even with large data
sets, whereas the naive approach has almost no power in all but the
simplest situations. Meanwhile, the latent variables approach worked
well in many examples, including an industry example presented in
Section~\ref{sechappyclickexample} that motivated this work.

\subsection{\texorpdfstring{Related work.}{Related work}}

Latent variables models have often been found to be powerful solutions
to weak supervision problems (also called distant supervision
problems). For example,
\citet{surdeanu2012multi} use a latent variables model to fit
distantly supervised relation extraction, and \citet
{tackstrom2011discovering} use a similar approach for sentence-level
sentiment analysis.

Generative structures similar to that underlying our latent variables
model have successfully been used in unsupervised topic modeling.
Prominent examples include probabilistic latent semantic analysis
[\citet{hofmann2001unsupervised}] and latent Dirichlet allocation
[\citet{blei2003latent}]. The idea of using weak or ambiguous
topic membership information to guide latent Dirichlet allocation has
been explored, among others, by \citet{toutanova2007bayesian} and
\citet{xu2009named}.

Other approaches to using side information in clustering include the
work of \citet{xing2002distance}, who showed how to enable
clustering algorithms to take into account user-provided examples of
similar and dissimilar pairs of points, and a group-wise support-vector
machine proposed by \citet{rueping2010svm}. \citet
{gordon1999classification} reviews methods for incorporating side
information into clustering algorithms using constraints.

\section{\texorpdfstring{Weakly supervised clustering.}{Weakly supervised clustering}}

Our goal is to cluster elements $i$ based on fine-grained features
$X_i$ in a way that aligns with side information on the average cluster
membership across various groups. Concretely, in the context of the
voting example, $X$ could encode voter demographic information $X = \{
\mbox{Income bracket, Union membership}, \ldots$\}, whereas in the web search
example, $X$ could be a click-level behavior $X = \{\mbox{Did the click
bounce}, \ldots$\}.

\begin{figure}

\includegraphics{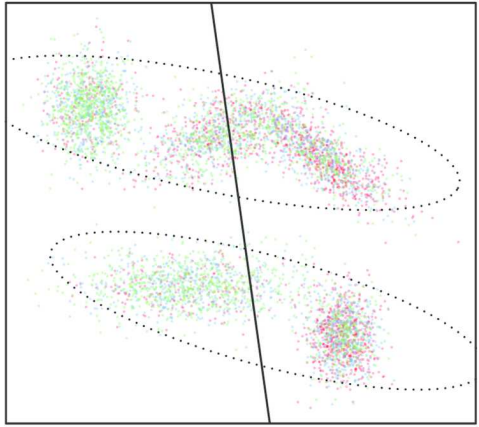}

\caption{A motivating example. Each dot corresponds to a single
click-level behavior. We know that dots corresponding to green, blue
and red dots are happy with probabilities of 80\%, 60\% and 10\%,
respectively. The data are drawn from a generative model for which the
solid line is the happy/sad decision boundary that minimizes logistic
loss. However, unsupervised Gaussian clustering divides the data into
two ellipses that are roughly orthogonal to the optimal decision
boundary. This paper develops methods that use side information about
the marginal happiness levels of green, blue and red clicks to help us
to recover the correct decision boundary.}
\label{figmotiv}
\end{figure}

To see the role of side information in weakly supervised clustering,
consider the following example. Suppose that, in the context of our web
search example, we have click-level data for 3 search results and that,
for visualization purposes, the click-level data $X$ can be represented
in 2 dimensions as in Figure~\ref{figmotiv}. Suppose, moreover, that
green and blue clicks are happy with probabilities of 80\% and 60\%,
respectively, but that red clicks are unhappy 90\% of the time; our
goal is to cluster these clicks into happy and sad clicks using this
side information.

%

If we did not have any side information, the best we could do is
attempt an unsupervised clustering of the data. Standard Gaussian
clustering as implemented in the \texttt{R}-library \texttt{mclust}
[\citet{fraley2012mclust}] divides the data into ellipsoids as
depicted in Figure~\ref{figmotiv}. It is quite clear that these
ellipsoids do not concur with our side information. In fact, given the
generative model used to produce the data, the best linear division of
our data into happy and sad clicks is given by the solid nearly
vertical line. Thus, the clustering obtained with unsupervised Gaussian
mixtures is roughly orthogonal to the division we would want.

An alternative baseline would be to ignore the latent structure of the
problem completely and simply set up a regression problem where we use
the coarse averaged labels as responses. Concretely, if we know that
80\% of green observations are happy, we could try to replace each one
of them with a positive example with weight 0.8 and a negative example
with weight 0.2. We discuss this approach further in Section~\ref
{secnaive}. In our experiments, this naive approach was unable to
capture most of the signal.

The goal of this paper is to develop techniques allowing us to use the
side information about the green, blue and red dots to recover the
decision boundary we want. Below we propose a generative model that
makes explicit the assumptions we need for weakly supervised clustering
to be possible. The subsequent section then proposes different ways of
fitting this generative model.

\subsection{\texorpdfstring{The key assumption.}{Weakly supervised clustering}}

The key assumption we need to make is that {click-level behavior is
conditionally independent of the side information given cluster
membership}. For example, in the context of our search provider
example, we assume that user behavior depends on satisfaction alone;
the search result quality only enters into the model through its
influence on user satisfaction.

This assumption can be represented using the graphical model in
Figure~\ref{figgraphsimple}. Let search results be indexed by $i \in
\{1,   \ldots,   I\}$ and clicks on the $i$th result be indexed by
$j \in\{1,   \ldots,   J_i\}$. Each result is associated with a
quality $\mu_i$, which affects whether individual clicks $j \in\{1,
  \ldots,   J_i\}$ on the result will be satisfied ($Z_{ij} = 1$) or
not ($Z_{ij} = 0$). The searcher then exhibits a click-level behavior
$X_{ij}$ that only depends on the satisfaction level $Z_{ij}$.

Our main assumption is that there is no edge going directly from $\mu$
to $X$. Thus, we force information to flow through the latent node $Z$
and thereby induce a clustering. A similar point is emphasized by
\citet{tackstrom2011semi}.

\subsection{\texorpdfstring{A generative model.}{A generative model}}

%
\begin{figure}[b]

\includegraphics{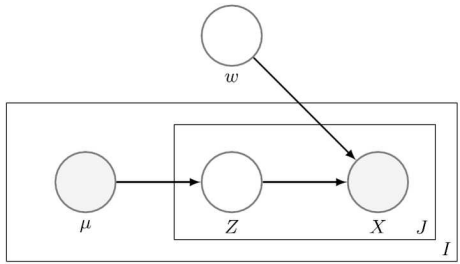}

\caption{Graphical model depicting the key assumption that $\mu_i$
and $X_{ij}$ are conditionally independent given $Z_{ij}$. Here, each
search result is associated with an underlying quality score $\mu_i$
which affects click-level user satisfaction $Z_{ij}$, which in term
influences behavior $X_{ij}$. The grayed-out nodes are observed, and
the boxes indicate repeated observations.}
\label{figgraphsimple}
\end{figure}

To build a practical weakly supervised clustering algorithm on top of
the conditional independence structure specified in Figure~\ref
{figgraphsimple}, we propose a simple generative model:
\begin{itemize}
\item Each search result $i \in\{1,   \ldots,   I\}$ has an
underlying quality $\mu_i \in\mathbb{R}$.
\item The satisfaction of each click $j \in\{1,   \ldots,   J_i\}$
on the $i$th search result is then independently drawn from the
Bernoulli distribution
\[
Z_{ij} \sim\operatorname{Bern}\bigl(\sigma(\mu_i)\bigr),
\]
where $\sigma(x) = 1/ (1 + e^{-x} )$ is the sigmoid function.
\item The searcher then exhibits a behavior $X_{ij} \in\{1,   \ldots,   K\}$ according to the multinomial distribution
\[
X_{ij} \sim\operatorname{Multinom}(w_{Z_{ij}}), %
\]
where\vspace*{1pt} $w_0$ and $w_1$ represent probability distributions on $\{1,
\ldots,   K\}$ (formally, they are vectors in $\mathbb{R}^K_+$ whose
entries sum to 1).
\end{itemize}
It is also possible to allow for more complicated distributional
assumptions for $X$: for example, $X_{ij}$ could be modeled as drawn
from a Gaussian mixture or from a cross-product of independent
multinomials. For our purposes, however, we found it simplest to
describe click-level behavior with a single binning obtained by
crossing multiple factors.

%
\begin{figure}[b]

\includegraphics{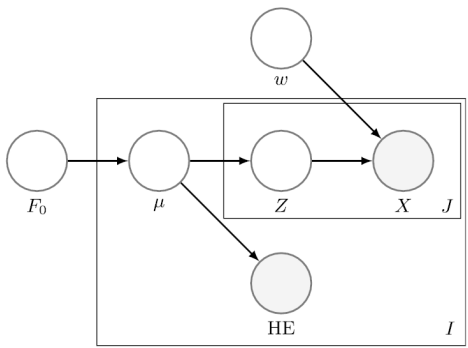}

\caption{Extension of the graphical model presented in Figure~\protect
\ref{figgraphsimple} that allows for the contingency that the $\mu
_i$ are not observed directly, but that we instead have noisy human
evaluation (HE) estimates of the $\mu_i$. The grayed-out nodes are
observed, and the boxes indicate repeated observations.}
\label{figgraphfull}
\end{figure}

In practice, we do not know the underlying quality $\mu_i$ and only
have noisy estimates of them. This is formalized in the graphical model
depicted in Figure~\ref{figgraphfull}. The true $\mu_i$ is drawn
from some prior $F_0$; we then get to observe a noisy estimate of $\mu
_i$ provided by outside human evaluation ($\HE$). For example, the
quantity $\HE$ could be obtained by asking workers on Amazon
Mechanical Turk to rate the likelihood that someone clicking on a
search result would be satisfied by it.
We model the rater noise as
$ \HE_i \sim\nn (\mu_i,   \sigma_H^2 )$.
The case with $\sigma_H^2 = 0$ reduces to the simpler model from
Figure~\ref{figgraphsimple}, while in the limit where $\sigma_h^2
\rightarrow\infty$ the outside information $\HE$ is only used for
initialization.

\subsection{\texorpdfstring{The estimand.}{The estimand}}

The key unknown parameters in our generative model are the multinomial
probabilities $w_0$ and $w_1$. From an interpretative point of view,
however, what we really want to know is the posterior probability that
a click was satisfied given a behavior. These can be obtained by Bayes' rule:
\[
\label{eqestimand} \rho{(k)}:= \PPp{Z = 1 | X = k} = \frac{\pi  w_1^{(k)}}{(1 -
\pi)   w_0^{(k)} + \pi  w_1^{(k)}},
\]
where the prior probability $\pi:= \PPp{Z = 1}$ is taken with respect
to the process that generated the $\mu_i$. We will frame all our
fitting procedures with the aim of estimating the posterior probability
vector $\rho$ instead of $w_0$ and $w_1$ themselves.

\section{\texorpdfstring{Simple baselines.}{Simple baselines}}

The hierarchical model defined in the previous section naturally lends
itself to being solved by maximum likelihood using an EM algorithm,
described in Section~\ref{secEM}.
That being said, the likelihood function of the whole latent variables
model is somewhat complicated and, in particular, is not convex. Before
going for a complex solution, we may want to check that simpler ones do
not work. In this section, we discuss some convex baselines. In the
experiments presented in Section~\ref{secexperiments}, we will find
the full maximum likelihood solution to vastly outperform its
competitors, suggesting that its complexity is not in vain.

\subsection{\texorpdfstring{A direct approach.}{A direct approach}}
\label{secnaive}

A first idea for dealing with the model in Figure~\ref
{figgraphsimple} is just to ignore the latent structure. Instead of
letting $Z_{ij}$ be a random Bernoulli variable with probability
parameter $\sigma(\mu_i)$, we could just create two artificial
observations: one with $Z_{ij} = 1$ and weight $\sigma(\mu_i)$, and
one with $Z_{ij} = 0$ and weight $1 - \sigma(\mu_i)$. In other words,
we swap out a single observation with an unknown latent label and
replace it with multiple observations with hard-assigned satisfaction
levels; the original probability parameter of the Bernoulli
distribution is used to set the weights of each artificial data point.
This transformation leads to simple estimates for the posterior
probabilities $\rho{(k)}$:
%
\begin{equation}
\label{eqnaive} \hrho{(k)} = \frac{1 + \sum_{i, j} \sigma(\mu_i)   1(\{X_{ij} =
k\})}{2 + \sum_{i, j} 1(\{X_{ij} = k\}) },
\end{equation}
where as usual we added one pseudo-observation in each behavioral bin
for numerical stability [e.g., \citet{agresti2002categorical}].

The main downside with this naive approach is that it cannot fit
variations in click-level behavior within groups and cannot account for
the fact that some clicks on bad search results may be happy and
vice-versa. As we will see in our examples, this will cost the method a
lot of power.
Ignoring the pseudo-observations, this naive approach is equivalent to
just training a linear regression with features $X_{ij}$ and response
$\sigma(\mu_i)$ (i.e., we regress the coarse responses on the fine
predictors directly). Thus, we can take the approach as a baseline for
what happens when we do not model latent click-level happiness.

\subsection{\texorpdfstring{Method of moments.}{Method of moments}}
\label{secmoments}

We can also try to estimate the $w_i$ by moment matching. If we set a
flat prior on the $\mu_i$ [or, equivalently a Haldane prior on $\sigma
(\mu_i)$] and provided that the number of replicates $J_i$ is
independent of $\mu_i$, then
\begin{eqnarray*}
\EEe{\sigma(\mu_i) | X_{i1}, \ldots, X_{iJ_i}} &=&
\frac{1}{J_i}\sum_{j = 1}^{J_i}
\rho{(X_{ij})}
\\
&=& \omega_i \cdot\rho,
\end{eqnarray*}
where $\omega_i$ is the empirical behavior distribution for the $i$th
search result: $\omega_i^{(k)} = |\{X_{ij} = k\}| / J_i$. Writing
$\sigma(\mu) \in[0,   1]^I$ for the vector containing the $\sigma
(\mu_i)$ and $\Omega$ for the matrix with rows $\omega_i$, we see that
%
\begin{equation}
\label{eqmean} \EEe{\sigma(\mu) | \{X_{ij}\}} = \Omega \rho.
\end{equation}
In practice, however, we know $\Omega$ and $\sigma(\mu)$, and want
to fit $\rho$ [with the model from Figure~\ref{figgraphfull}, we
can use $\sigma(\HE_i)$ as a surrogate for $\sigma(\mu_i)$]. We
could nevertheless try to use this moments equation as guidance and fit
$\rho$ by minimizing squared deviation from the moments equation
\eqref{eqmean}. This leads to an estimate
%
\begin{equation}
\label{eqMM} \hrho= \bigl(\Omega^\intercal\Omega\bigr)^{-1}
\Omega^\intercal\sigma(\mu).
\end{equation}
This estimator can perform well on very large data sets; moreover,
\citet{quadrianto2009estimating} establish theoretical regimes
where this method is guaranteed to perform well. However, we found it
to be prohibitively noisy on most of our problems of interest: the
estimates for $\hrho(k)$ are often not even contained in the $[0, 1]$
interval. The estimator also has some fairly surprising failure modes,
as discussed in Section~\ref{secmmnoise}.

\section{\texorpdfstring{An EM algorithm for the latent variables model.}{An EM algorithm for the latent variables model}}
\label{secEM}

In the previous section, we discussed some simple heuristic approaches
to weakly supervised clustering. Here, we show how to do maximum
likelihood estimation for the full latent variables model using an EM
algorithm [\citet{dempster1977maximum}]. The heuristic approaches
from before focused on the simpler model from Figure~\ref
{figgraphsimple}; EM, however, allows us the flexibility to work with
the full graphical structure from Figure~\ref{figgraphfull}. Our
likelihood function is not unimodal and so the proposed algorithm is
only guaranteed to converge to a local optimum rather than a global
one, but in practice our initialization scheme appears to have
consistently brought us near a good optimum. For a review of how the
EM-algorithm can be used to solve latent variables models see, for
example, \citet{bishop2006pattern}. Another algorithm that may be
worth considering for this problem is the MM-algorithm [e.g.,
\citet{lange2000optimization,hunter2004tutorial}].

All the individual steps taken by our EM-algorithm are simple and our
algorithm scales linearly in the size of the training data. Our
implementation in native \texttt{R} can handle around one million
clicks spread over ten thousand groups in just over 5 seconds.

\subsection*{\texorpdfstring{Initialization.}{Initialization}}
We initialize our model by forward-propagating the information obtained
from human evaluation (HE):
%
\begin{eqnarray}
\hmu_i &\leftarrow&\HE_i,
\\
\hz_{ij}&:=& \widehat{\PPp{Z_{ij} = 1}} \leftarrow\sigma(\hmu_i),
\\
\label{eqmstepw0} \hw_0^{(k)} &\leftarrow&\frac{1 + \sum_{ij}  (1 - \hz
_{ij} ) 1(\{X_{ij} = k\})}{K + \sum_{ij}  (1 - \hz
_{ij} )},
\\
\label{eqmstepw1} \hw_1^{(k)} &\leftarrow& \frac{1 + \sum_{ij} \hz_{ij} 1(\{X_{ij} =
k\})}{K + \sum_{ij} \hz_{ij}}.
\end{eqnarray}
We again added pseudo-observations for stability. This solution
effectively amounts to initializing our latent structure using the
naive model from Section~\ref{secnaive}.

\subsection*{\texorpdfstring{E-step.}{E-step}}
Given estimates for $\hmu_i$, $\hw_0$ and $\hw
_1$, the E-step for inferring latent variable probabilities $\hz_{ij}$ is
%
\begin{equation}
\label{eqestep} \hz_{ij} \leftarrow\frac{\sigma(\hmu_i) \cdot\hw_1^{(X_{ij})}}{
(1 - \sigma(\hmu_i)) \cdot\hw_0^{(X_{ij})} + \sigma(\hmu_i)
\cdot\hw_1^{(X_{ij})}}.
\end{equation}

\subsection*{\texorpdfstring{M-step.}{M-step}}
In the M-step, we need to update both the $\hmu$
and the $\hw$ given fixed estimates of $\hz$. The M-step for $\hw$
is the same update rule we used in our initialization, namely, (\ref
{eqmstepw0}), (\ref{eqmstepw1}).
Meanwhile, our updated estimate for $\hmu_i$ must maximize the
marginal log-likelihood, that is,
%
\begin{eqnarray}
\hmu_i &= &\argmin_{\mu_i} \Biggl\{\frac{(\mu_i - \HE_i)^2}{2
\sigma_H^2}
\nonumber\\[-8pt]\\[-8pt]\nonumber
&&\hspace*{46pt}{} - \sum_{j = 1}^{J_i} \bigl(
\mu_i \hz_{ij} - \log\bigl(1 + e^{\mu
_i}\bigr)
\bigr) + \log \bigl(f_0(\mu_i) \bigr) \Biggr\}.
\end{eqnarray}
For appropriate choices of prior density $f_0$, the minimization
objective is convex and the solution $\hmu_i$ is uniquely defined by a
first-order condition on the gradient. Putting an improper flat prior
on $\mu_i$, we get
%
\begin{equation}
\label{eqmstepmu} \frac{\hmu_i - \HE_i}{\sigma_H^2} + \sum_{j = 1}^{J_i}
\bigl(\sigma(\hmu_i) - \hz_{ij} \bigr) = 0.
\end{equation}
The left-hand side of the above expression is monotone increasing in
$\hmu_i$, and so this equation has a unique solution. We are not aware
of a closed-form solution to \eqref{eqmstepmu}; however, Newton's
method works well and is easy to implement for this problem.

\subsection{\texorpdfstring{Final answer.}{Final answer}} After iterating EM to convergence, we obtain
final estimates for the posterior probabilities
%
\begin{equation}
\label{eqlatentfinal} \hrho{(k)} = \frac{1 + \sum_{i, j} \hz_{ij}   1(\{X_{ij} = k\})}{2
+ \sum_{i, j} 1(\{X_{ij} = k\}) }.
\end{equation}

\subsection*{\texorpdfstring{A single tuning parameter.}{A single tuning parameter}} The only tuning parameter in the
update steps defined above is the noise variance $\sigma^2_H$ of the
human evaluation estimate $\HE_i$. As the form of \eqref
{eqmstepmu} makes clear, however, $\sigma_H^2$ only enters into the
model as a way to balance the relative importance of $\HE_i$ and the
$\hz_{ij}$ in estimating $\hmu_i$; thus, we expect our model to be
fairly robust to misspecification of this parameter. In our
experiments, we just used
$ \WH:= 1/\sigma_h^2 = 10$,
where $\WH$ stands for ``weight given to human evaluation.''

\subsection*{\texorpdfstring{Standard error estimates.}{Standard error estimates}}
In our experiments, we obtained
error bars for the parameter estimates by grouped subsampling: we
generated random subsamples by randomly selecting $I/2$ groups without
replacement and then looked at how much our point estimates varied when
trained on different subsamples. In general, half-sampling without
replacement is closely related to full sampling with replacement [e.g.,
\citet{efron1983estimating,politis1999subsampling}]. In this
problem, we chose to use subsampling instead of a nonparametric
bootstrap [\citet{efron1993introduction}] because we did not want
to have duplicate groups with identical click distributions.

\section{\texorpdfstring{Simulation experiments.}{Simulation experiments}}
\label{secsimu}

%
\begin{figure}[t]

\includegraphics{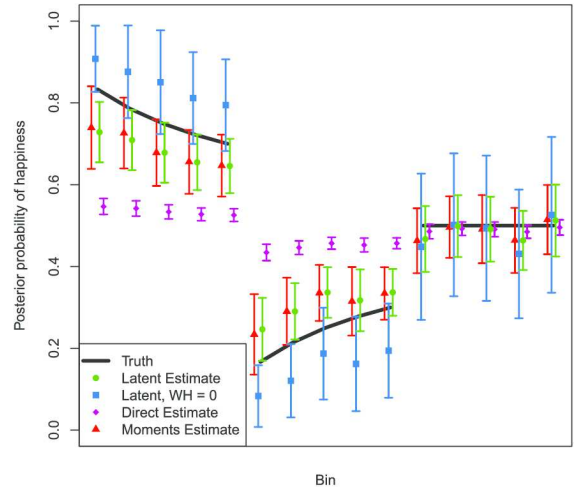}

\footnotesize{(a) $J_i = 5$ clicks per group}\vspace*{6pt}

\includegraphics{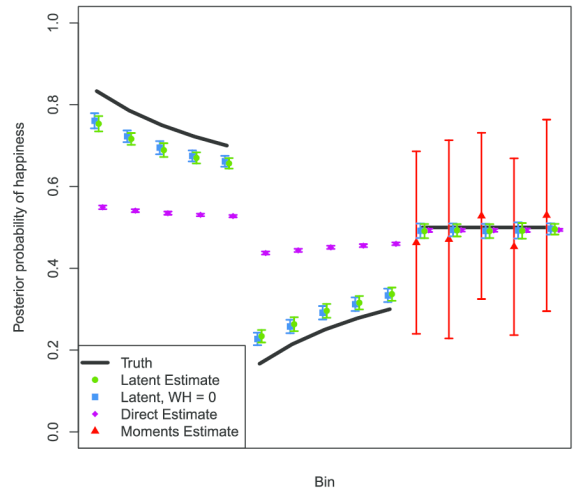}

\footnotesize{(b) $J_i = 100$ clicks per group}
\caption{Simulation example with many clicks per group. We have $I =
500$ groups with $J_i = 5$ or $100$ clicks each; behaviors are divided
into 15 bins with posterior probabilities of happiness given by the
thick black line. The data was generated with $\sigma_H = 0.5$; the
$\mu_i$ themselves were independently drawn from $\nn(0, 1)$. For the
latent variables model, we used $\WH= 10$. Error bars are 1 SD in each
direction and illustrate instability across 50 simulation runs.}
\label{figsimulations}
\end{figure}

We begin our empirical evaluation of weakly supervised clustering
methods with some simulation examples; Section~\ref{secexperiments}
has larger real-world examples.

The number of clicks per group can have a large impact on the relative
performance of different methods. In Figure~\ref{figsimulations}, we
show examples with $J_i = 5$ and $J_i = 100$ clicks per group. With 5
clicks per group, both the method of moments estimate and our latent
variables model perform reasonably well; the naive estimate that
directly hard-assigns cluster memberships underfits badly.

When there are relatively few clicks per group, weak supervision is
important: if we set $\WH= 0$ and only use the human evaluation data
for initialization, our latent variables model is prone to overfitting
and exaggerating the dynamic range of posterior probabilities. Using a
nonzero value of $\WH$ fixed this problem (we used $\WH= 10$).

The 100 clicks per group example looks quite different. First of all,
almost paradoxically, the method of moments estimate appears to have
gotten much worse as we added more data; estimates for the first 10
bins are not even contained in the $[0, 1]$ interval and so do not fit
into the plot. We propose an explanation for this surprising phenomenon
in Section~\ref{secmmnoise}.

Meanwhile, both latent variables procedures perform well. With many
clicks per group, the importance of the human evaluation $\HE_i$ after
initialization appears to fade away, and if we start off the EM
algorithm at a good spot, it can get itself to a desirable solution
without further guidance from the weak supervision; see Section~\ref
{secHEdisc} for more discussion.

\section{\texorpdfstring{Real-world experiments.}{Real-world experiments}}
\label{secexperiments}

In Section~\ref{sechappyclickexample}, we apply our method to the
problem that motivated our research: distinguishing satisfied from
unsatisfied clicks based on click-level behaviors. However, due to
confidentiality concerns, we need to present our results at a high
level and are not able to share details such as feature names.

To provide more insight into our method, we begin by presenting an
analysis of publicly available data from the 1984 presidential election
using our method. We assume a setting where we do not have access to
data on individual votes and need to rely on aggregated state-level
election results. Since the available labels are coarser than the
signal we want to fit, we need to do weakly supervised clustering to
learn about individual voter-level characteristics.
Although this application may appear quite different from the rest of
the examples we discuss, the underlying statistical task is very
similar. When constructing this example, we tried to make our analysis
mirror the analysis from Section~\ref{sechappyclickexample} as
closely as possible.

\subsection{\texorpdfstring{Weakly supervised clustering of voter demographics.}{Weakly supervised clustering of voter demographics}}
\label{secvoteexample}

In this example, we want to identify voter groups that favored the
Mondale/Ferraro ticket over Reagan/Bush in the 1984 US presidential
election and to build a model of the form
%
\begin{equation}
\label{eqvote} \PPp{\mbox{Vote for Reagan}} \sim f (\mbox{Demographic
Information} ).
\end{equation}
If we had access to a joint data set that records both individual votes
and individual demographic information, we could easily fit \eqref
{eqvote} by logistic regression. Here, however, we assume that we do
not have access to such a data set and that, for example, we only have
access to (1) a census data set with individual-level demographic
information that does not record voting intent, and to (2) state-level
aggregated election results. The problem of fitting \eqref{eqvote}
then becomes a weakly supervised clustering problem where, using
notation from Figure~\ref{figgraphsimple}, the $\mu$ represent
state-level election results, the $X$ are rows in the census data set,
and the $Z$ are inferred votes corresponding to the $X$.

More specifically, we base our analysis on a vector $\pi$ which
records the fraction of votes for Reagan in each state, and a design
matrix $X$ with the following per-voter information:
\begin{itemize}
\item State;
\item Annual income $\in\{1: [0,\$\mbox{12,500})$; $2: [\$\mbox{12,500},   \$
\mbox{25,000})$; $3: [\$\mbox{25,000}$,\break  $\$\mbox{35,000})$;
$4:[\$\mbox{35,000}, \$\mbox{50,000})$; $5: [\$\mbox{50,000}, \infty)\}$;
\item Union membership $\in\{$voter is a member of labor union, voter
has a family member who is a member of a labor union, voter has no
family members who are labor union members$\}$;
\item Race $\in\{$black, white, other$\}$.
\end{itemize}
The matrix $X$ has information on 8082 voters spread across 42 states,
with a median of 129.5 voters per state. The demographic factors have
the following frequencies: income $\{1342,  2331,  1897,  1473,
1039\}$, Union membership $\{1238,  1001,  5843\}$, and race $\{930,
  6869,   283\}$.

We constructed our data set based on election day exit-poll data
collected by CBS News and The New York Times following the 1984 US
presidential election, available from Roper Center for Public Opinion
Research at the University of Connecticut (USCBSNYT1984-NATELEC). We
removed entries for people who did not vote for either Mondale or
Reagan or who had missing data; before doing this, the original data
set had 9174 rows.\footnote{Of course, it would have been closer to
the spirit of our example to construct the data set $(X,   \pi)$
based on actual census data and aggregated voting information. For the
purpose of testing our methodology, however, using an exit poll data
set is advantageous: since we know what the actual votes were, we can
both check if our algorithm is making reasonable voter-level
predictions and compare its performance to an oracle model that gets to
use information about individual votes.}

Results are presented in Table~\ref{tabvote} for both the direct
method from Section~\ref{secnaive} and our latent variables approach.
In terms of cross-validation error, we see that the latent variables
method is almost on par with an oracle that gets to see individual
votes, whereas the direct method is not much better than just always
predicting the global mean. Note that for 0--1 error we did not tune the
decision threshold and just set it to even odds.\footnote{We do not
report results for the method of moments approach, as it did not work
at all here. In light of the examples from Section~\ref{secsimu},
this is not very surprising, as here we have few states and many voters
per state. For the latent variables method, we set $\WH= \infty$
because our state-level vote averages were accurate enough that we
could safely fix the $\mu_i$.}

\begin{figure}

\includegraphics{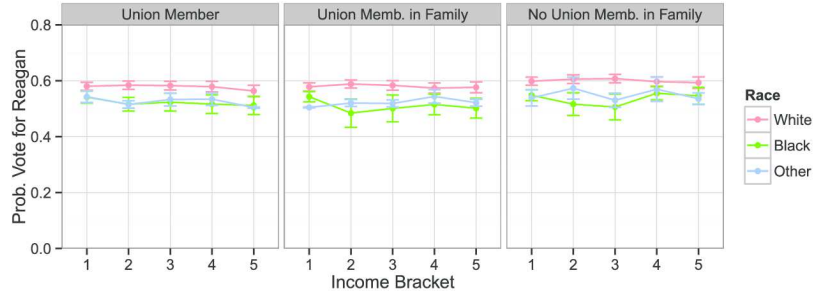}

\footnotesize{(a) Fit by direct hard assignment of labels}\vspace*{6pt}

\includegraphics{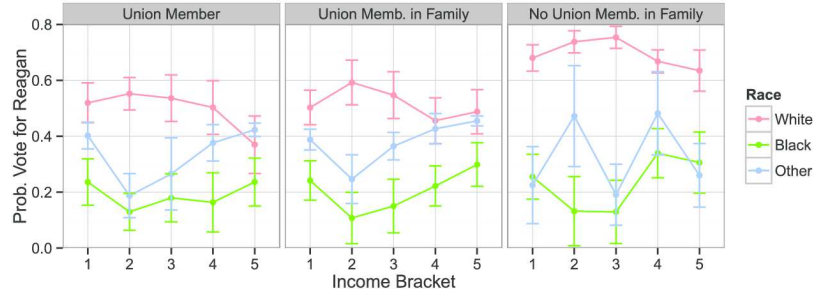}

\footnotesize{(b) Fit by latent variables modeling}\vspace*{6pt}

\includegraphics{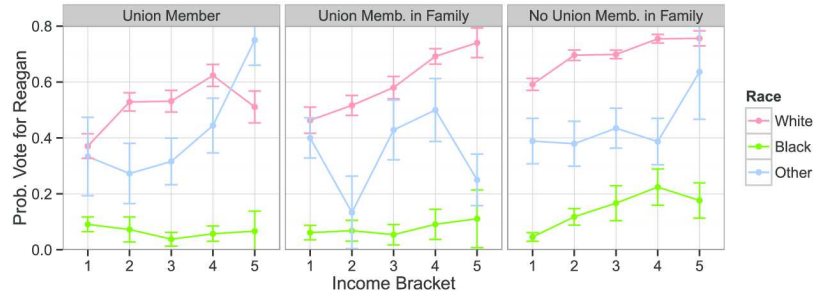}

\footnotesize{(c) Oracle fit}

\caption{Comparison of models fit by the direct method and the latent
variables method on the voter demographic example described in
Section~\protect\ref{secvoteexample}. In the last panel, we also
display the fit produced by an oracle that has access to the hidden
individual labels. All error bars are 1 SE in each direction and were
obtained by subsampling. We note that there are only 53 voters in the
$\mbox{``Union member} \times \mbox{Other race''}$ group and 29
voters in the $\mbox{``Union memb. in family} \times \mbox{Other
race''}$ group, so these two curves should not be interpreted too closely.}
\label{figvote}
\end{figure}

Figure~\ref{figvote} shows the predictions made by both the direct
and latent methods. We observe that the latent predictions have a much
wider dynamic range than the direct ones, which can be helpful if we
want to interpret the model predictions and get an intuition for effect
sizes. The latent variables predictions are also much more closer to
the gold-standard oracle predictions shown in the lowest panel. The
mean-squared difference between the latent variables model and the
oracle model, averaged over all 45 available factors, was 0.02; in
comparison, the mean-squared difference between the naive and oracle
models was 0.08.

\subsection{\texorpdfstring{Finding happy clicks.}{Finding happy clicks}}
\label{sechappyclickexample}

The research developed in this paper was motivated by a problem faced
by an internet company. In the terminology of our running example, we
had data on millions of click-level behaviors spread across thousands
of search results. We then asked a panel of annotators to estimate, for
each group, whether or not a click in a given group would likely lead
to satisfaction. Our goal was to learn to identify ``happy clicks''
based on click-level behavior; in other words, we wanted to perform a
weakly supervised clustering for click-level happiness.

%
\begin{table}
\tabcolsep=0pt
\caption{Results for predicting individual votes. The ``null model''
is the model-free baseline, which just guesses that every voter has the
same probability of voting for Reagan. The direct and latent models are
as described in Sections~\protect\ref{secnaive} and \protect\ref
{secEM}. The oracle model gets to see individual votes during
training; this is equivalent to training a direct model with a separate
group for each voter. With the exception of the null, all error rates
are cross-validated: we repeatedly trained each model on a random
sample of 21 states, and then evaluated the error rate on the remaining
21 states}\label{tabvote}
\begin{tabular*}{\tablewidth}{@{\extracolsep{\fill}}@{}lcccc@{}}
\hline
& \textbf{Null model} & \textbf{Direct} & \textbf{Latent} & \textbf{Oracle} \\
\hline
Mean classification error & 0.41 & 0.39 & 0.33 & 0.31 \\
Root mean squared error & 0.49 & 0.48 & 0.46 & 0.45 \\
\hline
\end{tabular*}
\end{table}

%

The distribution of clicks was heavily skewed. To avoid our result
being dominated by a few unusually large groups, we down-weighted
clicks in large groups such that the effective number of clicks in any
group was at most $M$, where $M \approx500$.
After down-weighting, the average number of clicks per group was around
one hundred.
Not down-weighting the biggest groups could lead to undesirable
consequences, as it could cause us to overfit to certain websites: for
example, if our training set contained a million clicks navigating to
\texttt{facebook.com} and we did not down-weight them, we might easily
overreact to special behaviors associated with Facebook clicks.

%
\begin{figure}[t]

\includegraphics{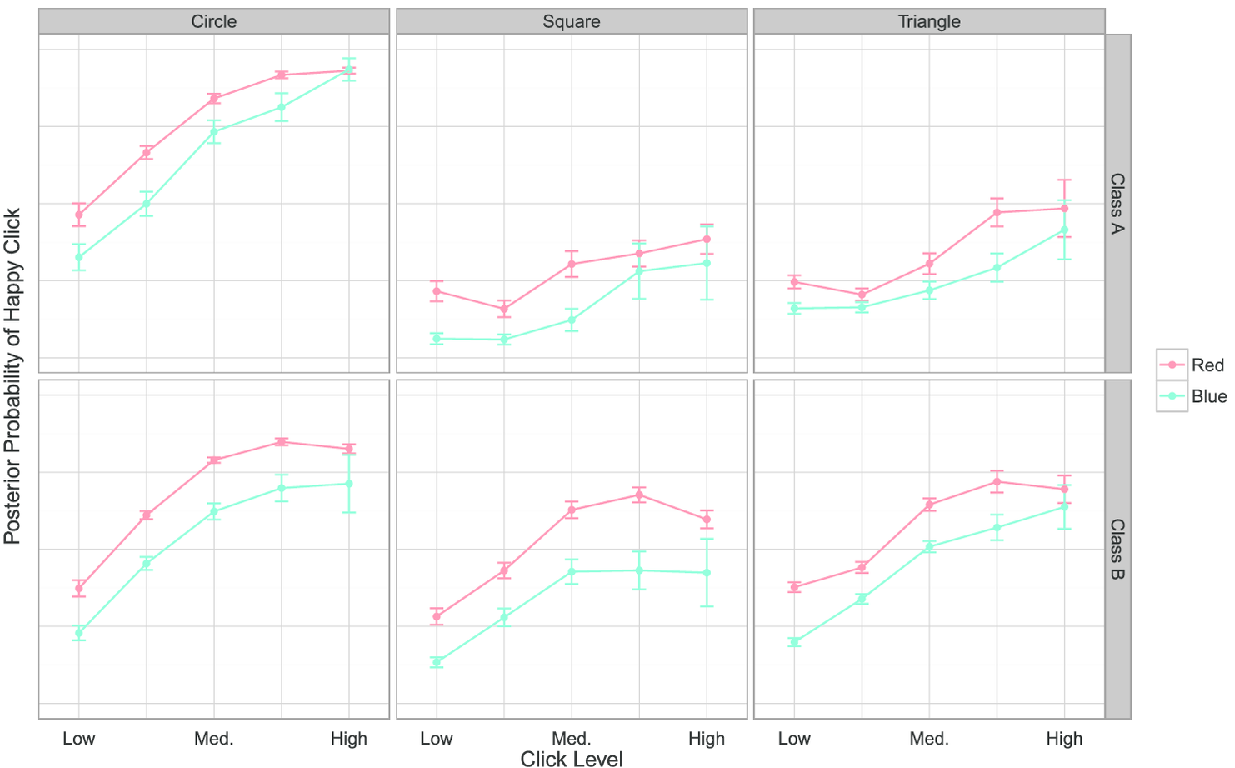}

\caption{Results of a real-world weakly supervised clustering analysis
described in Section~\protect\ref{sechappyclickexample}, using the
full latent variables model trained by EM. The groups are divided into
two classes (A and B) that were fit separately; click-level behaviors
are described by 30 buckets obtained by crossing level, shape and color
features (the true feature names have been obfuscated for
confidentiality reasons). All error bars are 1 SE in each direction and
were obtained by subsampling. We set the human evaluation tuning
parameter to $\WH= 10$.}
\label{figclicklatent}
\end{figure}

Results of our analysis are presented in Figure~\ref
{figclicklatent}. For confidentiality reasons, we cannot publish
feature names or axis scales.
The groups are split into two different classes (A and B), for which we
performed analysis separately. We described click-level behavior using
a full cross of three different factors, resulting in $5 \times3
\times2 = 30$ bins. Each point represented in Figure~\ref
{figclicklatent} was fit separately; the fact that these points seem
to fit along smooth curves suggests that our method is capturing a real
phenomenon.

%
\begin{figure}

\includegraphics{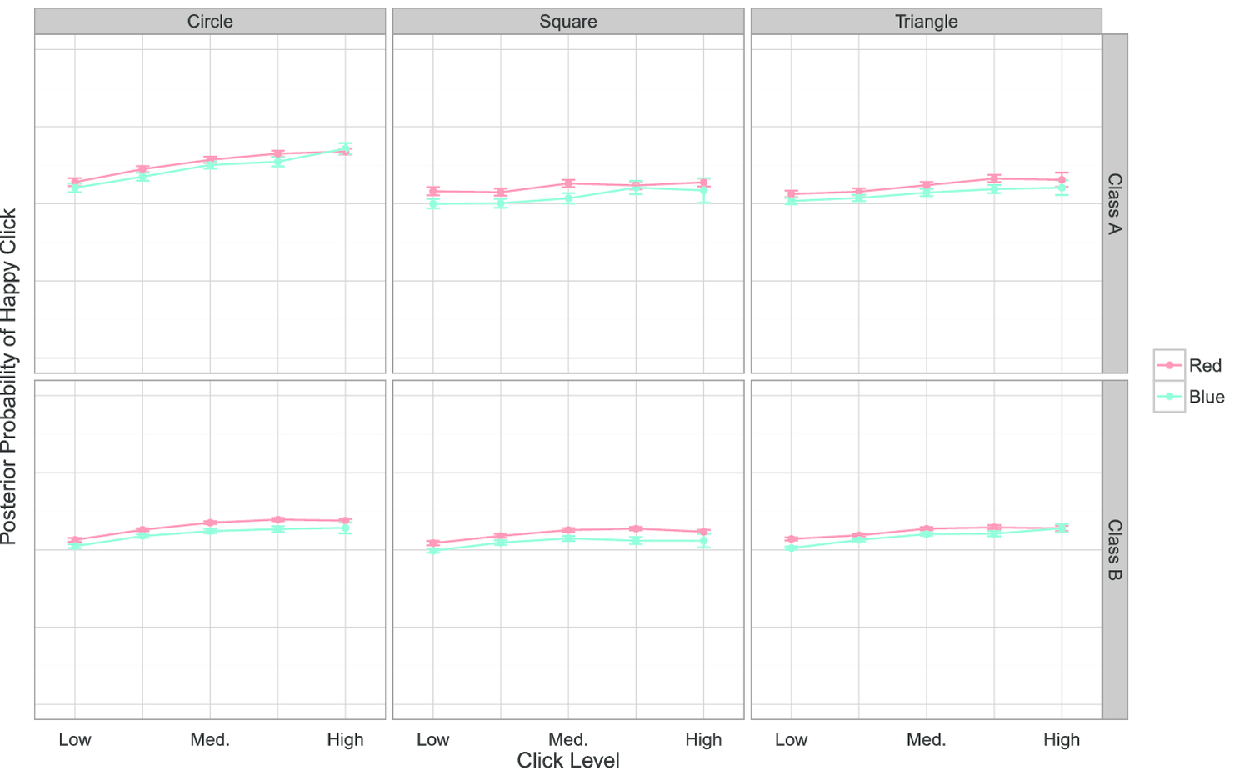}

\caption{Same analysis as that presented in Figure~\protect\ref
{figclicklatent}, except fit using the naive method from
Section~\protect\ref{secnaive}. The range of the $y$-axis is the
same as in Figure~\protect\ref{figclicklatent}; we see that the
naive method loses almost all of the dynamic range of the full model.}
\label{figclickdirect}
\end{figure}

%
\begin{figure}

\includegraphics{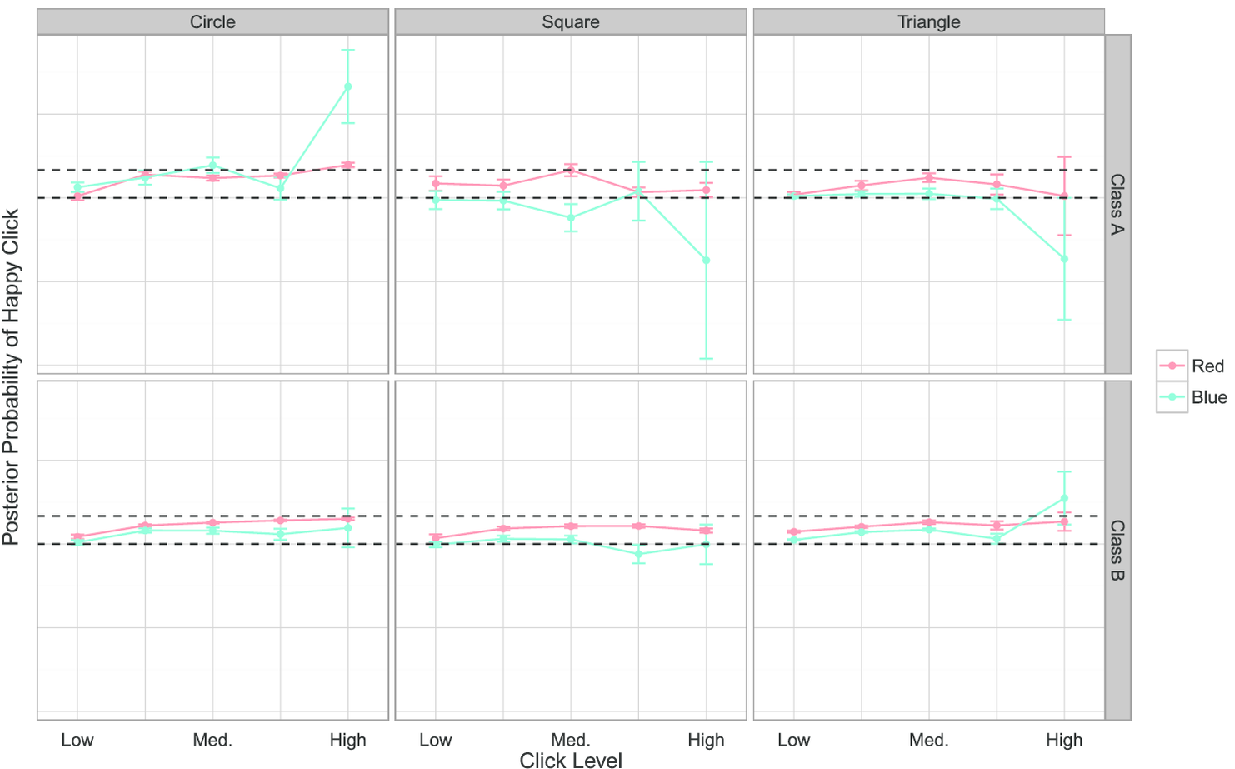}

\caption{Same analysis as that presented in Figure~\protect\ref
{figclicklatent}, except fit using the method of moments estimate
from Section~\protect\ref{secmoments}. The dashed lines indicate the
$y$-axis limits from Figure~\protect\ref{figclicklatent}. The
method of moments estimate appears to be severely unstable here.}
\label{figclickmoment}
\end{figure}

Latent variables modeling allowed us to discover multiple relationships
between click-level behavior and happiness, some of which confirmed our
intuitions and others which surprised us.
In terms of the obfuscated labels, we found that
(1)~Happiness generally increases with level, but with diminishing returns;
(2)~Red clicks are systematically more indicative of happiness than
blue clicks; and
(3)~Circular clicks are generally happier than square or triangular
ones, but this distinction is much more pronounced in Class A than in
Class B.
Of these facts, (1) was roughly expected and (2) had been conjectured,
although we were not expecting such a strong effect, but (3) came
largely as a surprise. We thought that Class A clicks should uniformly
be happier than Class B clicks, but it turns out that this relation
only holds for circles.

In Figures~\ref{figclickdirect} and \ref{figclickmoment},
provided at the end of the paper, we show the results of applying the
naive and method of moments estimates to this problem. These estimates,
respectively, under- and overfit the signal so badly that they did not
allow us to discover any of the key insights described above.


\section{\texorpdfstring{Discussion.}{Discussion}}

The simulation results from Section~\ref{secexperiments} suggested
some interesting relationships between the number of clicks per group
and the relative performance of various methods. Here, we present some
possible explanations for these relationships and also discuss
potential alternatives to our method.

\subsection{\texorpdfstring{The importance of human evaluation.}{The importance of human evaluation}}
\label{secHEdisc}

The human evaluation data $\{\HE_i\}$ enters into our EM-algorithm in
two locations: initialization, and the M-step for $\hmu_i$. Good
initialization is important, as it gives the algorithm guidance about
what kind of clustering to look for.
From our simulations, however, it appears that keeping $\HE_i$ around
for the M-steps is important when the number of clicks per group is
small, but less important when the number of clicks is large.

This phenomenon can be understood by looking at the M-step equation
\eqref{eqmstepmu}. We see that the relative importance of $\HE_i$
relative to the $\hz_{ij}$ in updating $\hmu_i$ scales inversely with
the number of clicks $J_i$ in group $i$. Thus, $\HE_i$ provides useful
support for updating $\hmu_i$ during the M-step when $J_i$ is small.
When $J_i$ is large the contribution of $\HE_i$ during the M-step gets
washed out, and our algorithm drifts more and more toward an
unsupervised clustering algorithm that uses human evaluation data for
initialization only. It appears that, in practice, with enough data per
group, human evaluation is only required to start the algorithm off
near the right mode.

\subsection{\texorpdfstring{Understanding the method of moments.}{Understanding the method of moments}}
\label{secmmnoise}

In our simulations, we found the method of moments estimate to perform
\emph{less} well as we added more clicks per group. Although this may
seem like a highly unintuitive result, we can attempt to understand it
using classical results about the connection between noisy features and
regularization.

The design matrix $\Omega$ used to fit the method of moments estimator
in \eqref{eqMM} records the fraction of clicks in each group that
appeared in a given bucket. The more clicks we have per group, the
closer each row of $\Omega$ gets to the true underlying behavior
distribution for each group. If the number of clicks per group is
small, then the rows of $\Omega$ are effectively contaminated by
mean-zero noise.

It is well known that training linear regression with a design matrix
corrupted by mean-zero noise is equivalent to training with a noiseless
design matrix and adding an appropriate ridge (or $L_2$) penalty to the
objective [\citet{bishop1995training}]; this connection between
noising and regularization has even been used to motivate new
$L_2$-like regularizers by emulating noising schemes [\citet
{maaten2013learning}, \citet{wager2013dropout}].

Now, if our model is correct, the noiseless limit of the rows of
$\Omega$ are in a \mbox{2-}dimensional space spanned by the happy and sad
behavior distributions. Thus, in the absence of noise, the regression
problem implied by our method of moments estimate is highly
ill-conditioned. But, when we only have few clicks per row, we are
effectively adding noise to $\Omega$ and this noise is acting as a
ridge penalty. Thus, for the method of moments estimator, throwing away
data can be seen as a (rather roundabout) way of fixing numerical
ill-conditioning.

\subsection{\texorpdfstring{Discriminative weakly supervised classification?}{Discriminative weakly supervised classification}}
\label{secdiscriminative}

For our latent variables approach, we chose to treat $X$ as a random
variable depending on $Z$ and to model $\law(X | Z)$. An alternative
choice would be to set up a discriminative model where we condition on
$X$ and model $\law(Z|X)$; in terms of the plate diagram from
Figure~\ref{figgraphsimple}, this would amount to swapping the
direction of the arrow from $Z$ to $X$.

This class of models has been studied in detail in the context of
logistic regression with unreliable class labels [e.g., \citet
{copas1988binary,magder1997logistic,yasui2004partially,kuck2005learning}]:
given a data set of $(X,   Y)$-pairs with $Y \in\{0,   1\}$, the
authors posit that the observed class labels $Y$ are potentially
erroneous, but that there exist unobserved true labels $Z \in\{0,
1\}$ such that $(X,  Z)$ are drawn from a logistic regression model
and $\PPp{Y = Z} = 1 - \varepsilon$. Formally, this results in a
probabilistic model where $\law(Z |X) = \operatorname
{Bernoulli}(\sigma(\beta\cdot X))$ for some parameter vector $\beta
$, and then $\law(Y | Z) = \operatorname{Bernoulli}(\varepsilon+ Z
  (1 - 2\varepsilon))$.

The main difference between the noisy class labels problem and our
problem is that the former has a natural model for $\law(Y | Z)$,
whereas in our setup $\mu$ does not depend causally on $Z$. In our
motivating examples we think of $\sigma(\mu)$ as (a potentially noisy
estimate of) the \emph{population} mean of the $Z$, such that $\mu$
is conditionally independent of $Z$ given the population. Thus, our
problem statement does not fit directly into the framework of
\citet{copas1988binary} and others.

\section{\texorpdfstring{Conclusion.}{Conclusion}}

Classification problems where training labels are much coarser-grained
than the signal we are trying to fit arise naturally in many
applications. We showed how they can be formalized as weakly supervised
clustering problems and presented three approaches to fitting them,
including a latent variables model that worked well in our experiments.
In both the elections application from Section~\ref{secvoteexample}
and the real-world problem that originally motivated our research
(Section~\ref{sechappyclickexample}), our method enabled us to gain
qualitatively richer insights than baselines that rely on hard
assignment of labels or moment matching. An interesting topic for
further work would be to study the information loss from only having
access to weakly instead of fully labeled data.

\section*{\texorpdfstring{Acknowledgment.}{Acknowledgment}}
The authors are grateful to Nick Chamandy,
Henning Hohnhold, Omkar Muralidharan, Amir Najmi, Deirdre O'Brien, Wael
Salloum, Julie Tibshirani and Brad Efron for many helpful discussions,
and to their AOAS editor, Brendan Murphy, for constructive feedback on
earlier versions of this manuscript.




%

\printaddresses
\end{document}